\documentclass[conference]{IEEEtran}
\IEEEoverridecommandlockouts
\usepackage{cite}
\usepackage{amsmath,amssymb,amsfonts}
\usepackage{graphicx}
\usepackage{comment}
\usepackage{textcomp}
\usepackage{xcolor}
\usepackage{color}
\usepackage{lipsum}
\usepackage{times}
\usepackage{textcomp}
\usepackage{amsmath,amssymb,amsfonts}
\usepackage[inline]{enumitem}
\usepackage{algorithm}
\usepackage{algpseudocode}
\usepackage{amsmath}
\usepackage{color}
\usepackage{tablefootnote}
\usepackage{mathtools}

\usepackage{amsthm}
\usepackage{imakeidx}
\usepackage{float}
\usepackage{multicol}
\usepackage{booktabs} % For formal tables
\usepackage{subcaption}
\usepackage{listings}
\usepackage{verbatim}
\usepackage{xcolor,colortbl}

\usepackage{multirow}
\usepackage{lscape}
\usepackage{verbatim}

\def\BibTeX{{\rm B\kern-.05em{\sc i\kern-.025em b}\kern-.08em
    T\kern-.1667em\lower.7ex\hbox{E}\kern-.125emX}}
\begin{document}

\title{
TriplePlay: Enhancing Federated Learning with CLIP for Non-IID Data and Resource Efficiency
}

\author{\IEEEauthorblockN{Ahmed Imteaj$^{1,2}$, Md Zarif Hossain$^{1,2}$, Saika Zaman$^{1,2}$, Abdur R. Shahid$^{1}$}
\IEEEauthorblockA{\textit{$^{1}$School of Computing, Southern Illinois University, Carbondale, IL, USA} \\
\textit{$^{2}$Security, Privacy and Intelligence for Edge Devices Laboratory (SPEED Lab)}\\
mdzarif.hossain@siu.edu, imteaj@cs.siu.edu, saika.zaman@siu.edu, shahid@cs.siu.edu} 
% \vspace{-1.02cm}
}
% Personalizing, Balancing and Streamlining Foundation Models for Federated Learning

% \author{\IEEEauthorblockN{Saika Zaman, Ahmed Imteaj
%     }
% \IEEEauthorblockA{\ School of Computing, Southern Illinois University, Carbondale, IL, 62901 \\
% saika.zaman@siu.edu, \{imteaj\}@cs.siu.edu}
% }

\maketitle

\begin{abstract}
The recent advancement of pretrained models shows great potential as well as challenges for privacy-preserving distributed machine learning technique called Federated Learning (FL). With the growing demands of foundation models, it is now an urgent need to explore the potential of such foundation models in a distributed setting. In this paper,  In this paper, we delve into the complexities of leveraging foundation models, like CLIP into FL frameworks to preserve data privacy, and efficiently training distributed network clients across heterogeneous data landscapes. We specifically aim to address the issues related to non-IID data distributions, skewed class representation of FL clients' local dataset, communication overhead and high resource consumption due to large, complex model training  in an FL setting. To address these, we propose TriplePlay, a framework that tailors CLIP foundation model as an adapter to strengthen FL model's performance and adaptability across heterogeneous data distributions among the clients. Besides, we address the long-tail distribution problem in an FL environment to maintain fairness and optimize the computational resource demands of the FL clients through quantization and low-rank adaptation techniques. A comprehensive simulations results with two distinct datasets and different FL settings demonstrate that TriplePlay efficiently reduces GPU usage and accelerates the convergence time that ultimately reduces the communication cost.
\end{abstract}

\begin{IEEEkeywords}
Federated Learning, foundation model, CLIP, personalization, GAN, resource optimization.
\end{IEEEkeywords}

\section{Introduction}
\subsection{Motivation}
CLIP (Contrastive Language-Image Pre-Training) \cite{radford2021learning} was developed by OpenAI, which is basically a neural network based model that is trained on a comprehensive and diverse dataset capturing images paired with corresponding textual details. This pre-training with an extensive dataset enables CLIP model to efficiently align visual data with relevant natural textual data. In consequence, such models have large size and knowledge capacity.
As FL cements its role as a key enabler of privacy-preserving artificial intelligence, it confronts challenges that stem from the inherent heterogeneity of data distributions across clients and the considerable resource demands of incorporating large foundation models like CLIP. These challenges, notably non-IID data and the substantial computational and communication costs, impede the seamless deployment and efficacy of FL in diverse operational environments characterized by a wide range of edge device capabilities. Despite its reputation for state-of-the-art image and text representations, CLIP remains relatively unexplored in FL, with prior efforts attempted to achieve efficient aggregation and local training but falling short in addressing computational costs and data distribution heterogeneity. Besides, in the realm of FL, the significance of data quality and distribution cannot be overstated as many datasets suffer from a common issue known as the `long-tail distribution', where certain classes or categories are underrepresented, making it challenging for models to learn effectively from these minority samples. Besides, large model can cost high network bandwidth and encounter communication overhead. Considering these challenges, this paper investigates the impact of integrating a foundation model like CLIP as an adapter in FL on the adaptability and performance of FL systems across varied data distributions, to transcend existing limitations and fully exploit the potential of this powerful model.
However, the existing foundation models face challenges in FL settings, specially when there are challenges related to data quality and certain class are underrepresented by a significant number that complicate the learning process of the models. This underscores the necessity of an effective solution to guarantee balanced and efficient learning of the FL clients. 

This paper tailors the robust feature extraction capabilities of CLIP and addresses the resource-constrained issues inherent in FL-based systems. Besides, we explore the quantization with low-rank adaptation technique to reduce the computation resource usage of the FL clients. This results in reducing the local model size and efficient model exchange with the FL server. To this end, this paper presents a privacy-preserving, scalable, and resource-efficient learning scheme that leverages the power of pretrained foundation models along with synthetic data generation to eliminate the issue of long-tail distribution for personalized and equitable AI applications.
% \vspace{-0.1cm}
\section{Background Study}

\subsection{Federated Learning}
Federated Learning (FL) \cite{mcmahan2017communication} is a distributed machine learning technique that allows multiple devices or institutions to collaboratively train a model without sharing their raw data. In the FL process, instead of sharing raw sensitive data, each participated client performs on-device model training independently on its local dataset and share the model with the FL server. The shared models are then aggregated on the FL server that produces a global model that captures all the clients local knowledge, while preserving privacy. Besides, the FL process handles a crucial challenge that we observe in real-world distributed setting, which is non-IID (non-independent and identically distributed) data. A distributed client could possess data which is different than the other distributed agents within the network. FL addresses that issue by collecting each client's local model that is trained on unique local data distribution and subsequently performing weighted average. The weighted average of the local models make sure to give more importance to the clients' model which has high data relevance. This process eventually results in an improved version of the global model in a privacy-preserving manner. FL shows its potential to a wide-range of domains where privacy and data ownership has utmost importance, e.g., healthcare, smart city, finance, and mobile applications \cite{nguyen2021federated, imteaj2021survey, hossain2024flamingo, 9356216}. However, FL faces various implementation challenges due to the data heterogeneity issues that force the clients' to generate skewed and diverge local model updates. In recent years, several efforts \cite{chen2022bridging, gupta2022fl, qu2022generalized, hossain2024fedavo, foret2020sharpness}
has been made to address to tailor the domain generalization methodologies to FL settings. The authors of FL Games \cite{gupta2022fl} proposed Nash equilibrium, while Qu et al. \cite{qu2022generalized} introduced sharpness aware minimization technique to promote consistent feature learning among clients and improve the overall robustness and generalization of the model. However, such generalization strategies has limited applicability to real-world FL settings with larger models \cite{li2021ditto, tan2022towards} and in pre-trained models \cite{tan2022federated, tian2022fedbert, chen2022importance, guo2023promptfl}. This gap highlights key challenges within the FL domain, indicating a pressing need for a novel approach to enhance generalization ability of FL with large models or large pre-trained models.

\subsection{Pretrained Vision-Language Models}
The pretrained Vision-Language Models (VLMs) are one type of large language models which are pretrained with extensive and diverse datasets consisting of visual images and their corresponding textual descriptions. One of the significant features of VLM is its capability of zero-shot task, where the VLM can classify or infer information from images that it never been exposed to. One such state-of-the-art model is CLIP \cite{radford2021learning}, which uses embedded image and text encoders to perform image classification or information retrieval. Another model named ALIGN \cite{jia2021scaling}, perform training on visual and text representation tailoring large image samples paired with noisy alt-text data. Besides, BLIP \cite{li2022blip} is another VLM that can generate vision-language content focusing on the refined captions. FLAVA \cite{singh2022flava} proposed a strategy to learn from both paired and unpaired visual and textual data and combines both unimodal and multimodal encoders for a comprehensive representation. Moreover, SimVLM \cite{wang2021simvlm} aims to reduce the training complexity by applying a large-scale weak supervision with a prefix language modeling. Further, several efforts are seen that extended the CLIP model to support multilingual text encoding, such as AltCLIP \cite{chen2023altclip}, and domain-specific adaptations like FashionCLIP \cite{chia2022contrastive} and PLIP \cite{huang2023visual}. VLMs has the potential to adapt considering various datasets and objectives, such as VLMs can be used in various applications, including domain-specific visual embedding generation \cite{wang2023fashionvqa, du2024domain, wang2024grammar, lian2024recai, gouidis2024fusing, li2023igg}, semantic segmentation \cite{sun2024training, wang2024llm, wei2024lasagna, cha2023learning}, object detection through knowledge distillation \cite{li2023object, park2024localized, 10633382, kang2024knowledge, imteaj2023fedmdp}, robustness and security \cite{hossain2024securing, hossain2024sim}, and the adaptation of CLIP for personalized supervised learning \cite{guo2023pfedprompt, yeh2023meta, cho2023promptstyler, yan2023clip, chen2024feddat}. This evolution and rapid advancements of pretrained VLMs paves the way to develop multimodal AI systems that can understand and generate human-life perceptions with the synergy between visual and textual data.

% This evolution of pretrained VLMs showcases the dynamic synergy between visual and textual data and paves the way for more efficient, and versatile AI systems capable of understanding and generating human-like perceptions of the world.

\begin{figure*}[htb!] % Defines figure environment
 \setlength{\belowcaptionskip}{-10pt}
    \centering % Centers your figure
    \includegraphics[width=0.9\linewidth]{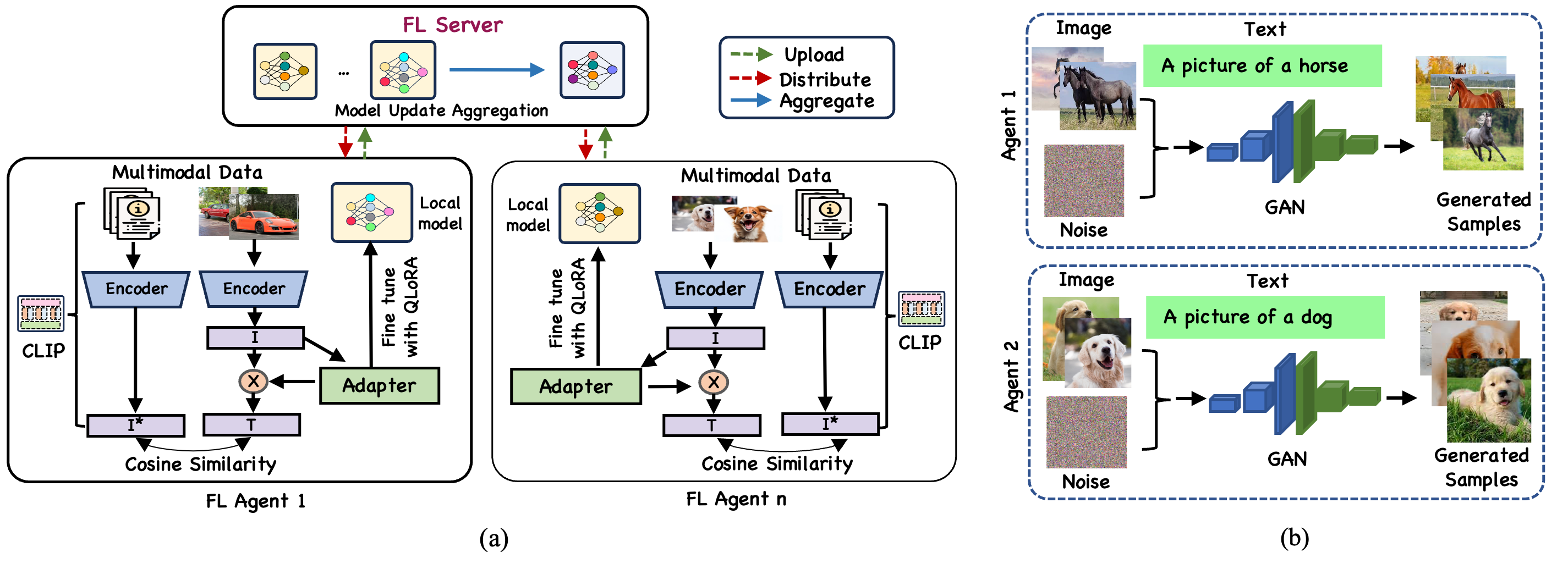} % Includes your figure and defines the size
    % \vspace{-0.4cm}
    \caption{(a) Overview of the TriplePlay system architecture, (b) FL client-sourced and GAN-generated image samples.}  % For your caption
    \label{fig:1}
\end{figure*}

\section{Proposed Approach}
Our proposed approach, TriplePlay focuses on three key tasks: personalization and generalization, handle underrepresented classes, and minimizing resource consumption, which are detailed below:
% \subsection{Prioritizing Personalization and Generalization:} Pretrained models inherently possess the capability to extract robust and diverse features. However, fine-tuning entire networks with limited data may compromise these models' original abilities. Therefore, our approach emphasizes preserving valuable prior knowledge and utilizing it optimally for specific tasks. Additionally, given the impracticality of tuning large networks in FL settings due to resource constraints, we focus on a simple attention-based adapter, seamlessly integrated with the CLIP model, facilitating rapid task-specific adaptation with minimal resource overhead. These adapters leverage CLIP's extensive pretrained knowledge base, tailored to meet the unique operational demands of FL environments. Additionally, this approach ensures that models maintain a high level of generalization capability, ensuring robust performance across unseen data and tasks.
\subsection{Prioritizing Personalization and Generalization:} To prioritize personalization and generalization in federated learning (FL), we propose a strategy that preserves valuable prior knowledge from pretrained models while adapting them to specific tasks efficiently. Fine-tuning entire networks with limited data can compromise their original abilities, specially in resource constrained FL settings. Considering that, we focus on an attention-based adapter technique with the integration of the CLIP model. This approach enables swift task-specific adaptation and a minimal overhead of resources. The steps are detailed below:

% To prioritize personalization and generalization in federated learning (FL) settings, we propose the following approach using a simple attention-based adapter integrated with the CLIP model:

\subsubsection{Pretrained CLIP Model} Initially, we start with a pretrained foundation model, CLIP denoted as \( \text{CLIP}_{\text{pre}} \). This model is pretrained on a diverse and large corpus of data.
\subsubsection{Adapter Architecture} Next, we design an attention-based adapter and added that on top of the \( \text{CLIP}_{\text{pre}} \). This mechanism helps the model to adapt in specific tasks. The adapter has two major components:
   
   \textit{Attention Mechanism:} This mechanism allows the adapter to simultaneously focus on different segments of the input data ($\mathcal{D}$) that enables task-specific adaptation:
   $$
\text{ Att}(\text{$\mathcal{D}$})=\operatorname{softmax}\left(Q \cdot K^T\right) \cdot V
$$
   where \( Q \), \( K \), and \( V \) are the query, key, and value matrices, respectively, derived from the input data.

   \textit{Feedforward Network:} After leveraging the attention mechanism, the output is feed into a feedforward network to further adapt the features. We represented the feedforward network as follows:
   $$
\operatorname{F_{net}}(\operatorname{Att}(\text{$\mathcal{D}$}))=\operatorname{ReLU}\left(W_1 \cdot \operatorname{Att}(\text{$\mathcal{D}$})+b_1\right) \cdot W_2+b_2
$$
   where \( W_1 \), \( b_1 \), \( W_2 \), and \( b_2 \) are the weights and biases of the feedforward network.

\subsubsection{Adapter Integration} Then, as an additional layer, we integrate the adapter into the CLIP model. We represented the adapted CLIP model as follows:
$$
\operatorname{CLIP}_{\text {adapted }}(\text{$\mathcal{D}$})=\operatorname{Adapter}\left(\operatorname{CLIP}_{\text {pre }}(\text{$\mathcal{D}$})\right)
$$

\subsubsection{Training Procedure} We perform training of the adapter using some task-specific data and assigning the weights of \( \text{CLIP}_{\text{pre}} \) as constant. During the training of the adapter, the goal was to minimize a task-specific loss function. For example, in classification tasks, we use cross-entropy loss and in regression tasks, we use mean squared error.

Extracting the pretrained features from \( \text{CLIP}_{\text{pre}} \) with the adapter assists the model to enhance the generalization capacity. Besides, the task-specific adaptation allows the adapter to allow personalization to specific tasks. This technique strikes a balance between generalization and personalization which is suitable for resource-constrained FL environment because the limited resources can hinder extensive fine-tuning.

\subsection{Handling Long-Tail Distribution} 
To handle long-trail distribution in an FL environment, we generate synthetic data using GANs which mitigates the data quality issue and also rebalanced the class distributions. For an imbalanced dataset, where a certain class of data samples are underrepresented comaapre to the rest of the other classes in the dataset, GANs can be helpful to generate synthetic data and mitigate the class imbalance issue. In Fig. \ref{fig:1}(b), we illustrate how GAN can generate various image samples for a particular visual and textual description. GAN generally perform augmentation of the dataset image samples, and generate various version of a sample visual data. This eliminates the challenge with long-tail distribution and allow the FL client to effectively learn even from the underrepresented class of image samples. This approach results in enhanced model generalization and fairness in model predictions across all classes. 

To perform training of the GANs, we consider two major components of GANs: Generator (\(G\)) and Discriminator (\(D\)). \(G\) produces data that is indistinguishable from the real data samples of the FL clients, while the task of the \(D\) is to distinguish the generated data by \(G\) and the real data.

\subsubsection{Discriminator's Goal} The discriminator, \(D\), aims to assign the correct labels to both real and generated data. It maximizes the probability of assigning the correct label to both real data (coming from the dataset) and fake data (produced by \(G\)). 
In GANs model training, we consider expected log probability, \(\mathbb{E}_{x \sim p_{data}(x)}[\log D(x)]\) that the discriminator correctly classified real data as real. The goal is to maximize the expected log probability value that indicates that the discriminator is reliable at recognizing real data.
    
\subsubsection{Generator's Goal} The goal of the generator \(G\) is to produce data fake data that will be incorrectly classified by the discriminator \(D\) as real. In GANs model training, we consider expected log probability, \(\mathbb{E}_{z \sim p_z(z)}[\log(1 - D(G(z)))]\) that mistakenly classifies fake data as real. The generator tries to minimize this expected log probability so that the discriminator believes the fake generated data as real.

    \subsubsection{The Min-Max Game} The min-max formulation \(\min_G \max_D V(D, G)\) captures the adversarial nature of the training process. \(D\) maximizes \(V(D, G)\) by getting better at distinguishing real from fake, while \(G\) minimizes \(V(D, G)\) by improving its ability to generate data that appears real to \(D\).

\vspace{0.2cm}
\textbf{Explanation:}
\begin{itemize}
    \item The discriminator's optimization (\(\max_D\)) increases its accuracy in distinguishing real data from fake. In this process, the goal is to maximize the probability to correctly classifying real and fake data.
    \item During the process of the generator's optimization (\(\min_G\)), the target is to deceive the discriminator by producing data samples which are imperceptible from real data. The goal is to minimize the discriminator's capacity to correctly labeling the fake data as fake. 
    \item Such adversarial process helps the generator \(G\) to produce more realistic data, and also assist the discriminator \(D\) to perform better labeling of fake and real data. 
\end{itemize}

This process continues until the data generated by \(G\) are indistinguishable from real samples, which means that the generator is now learned to successfully generate data that resembles the real data.

% This iterative training process continues until a point of equilibrium is reached where \(D\) can no longer distinguish between real and generated data, meaning \(G\) has successfully learned to generate data resembling the real data distribution.

% One promising approach to address the issue of underrepresented classes is the generation of synthetic data using Generative Adversarial Networks (GANs). We plan to train GANs to generate synthetic data samples that are statistically similar to real data, effectively augmenting the dataset size for rare classes and balancing the data distribution (see Figure \ref{fig:1}(b)). This augmentation effectively increases the dataset size for rare classes, helping to balance the data distribution. Integrating GAN-generated synthetic data into the FL training process can enhance model performance across all classes, including those in the long tail. The augmented datasets ensure that the model has sufficient examples from each class to learn from, leading to better generalization and a more equitable performance across the dataset's spectrum.
\subsection{Optimizing Resource Consumption and Reducing Communication Cost:} In the traditional foundation models for the distributed system, there are dual challenges: high resource consumption for training foundation models at the edges and extra communication overhead due to the large size model exchange between the FL server and the clients. We optimize the resource consumption and shrink the large model size applying quantization and QLoRa. Quantization shrinks the local model size of the FL clients by transforming parameters into lower-bit representations, that results in lowering memory usage and accelerating the computational speed. QLoRa improves the training process of quantized models of FL clients with an aim to minimize the model loss and improving model accuracy through an efficient learning from the local data.

Incorporating GAN-based synthetic data generation into the initial stages of feature extraction and adaptation, alongside the integration of QLoRa and quantization strategies for optimization, we redefine the methodology as follows:

\subsubsection{Advanced Feature Extraction with Synthetic Data Augmentation}
For each participating FL agent, a pretrained CLIP model is used at the outset to extract relevant features from textual and image data.
Given an input-label pair \((\mathbf{v}, z)\), enhanced with GAN-generated synthetic data to address class imbalance and data underrepresentation, we utilize the pretrained CLIP model for comprehensive feature extraction:
% \vspace{-0.2cm}
$$
\mathbf{V}_{\textit{st}} = f_{\text{vis}}(\text{GAN}(\mathbf{v})), \quad \mathbf{U}_{\textit{st}} = f_{\text{text}}(\text{GAN}(z))
$$

Here, \(\text{GAN}(\mathbf{v})\) and \(\text{GAN}(z)\) represent the synthetic visual and textual data generated to enrich the training dataset, ensuring a more balanced and diverse feature set for the learning process.

\subsubsection{Task-specific Feature Refinement}
An adaptive feature refinement method \(h\) that can selects crucial feature for domain specific tasks and can be adaptively refined when more synthetic data are generated. We integrate an attention mechanism to the original and synthetic features to focus on most relevant information, which can be presented as follows:

% An adaptive refinement mechanism \(h\), further informed by the diversity and balance brought in by synthetic data, selectively enhances features crucial for task-specific needs. This process involves an attention mechanism applied to both original and synthetic features:
% \vspace{-0.2cm}

$$
\mathbf{V}^{'} = h(\mathbf{V}_{\textit{st}}) \otimes \mathbf{V}_{\textit{st}}, \quad \mathbf{U}^{'} = h(\mathbf{U}_{\textit{st}}) \otimes \mathbf{U}_{\textit{st}}
$$

\subsubsection{Feature Normalization and Interaction with QLoRa and Quantization}

After the feature refinement, we pass the features through a normalization process tailoring QLoRa and quantization techniques for optimized computation and interaction:

$$
\mathbf{V}_{\text{opt}} = \frac{\mathbf{V}^{'}}{\|\mathbf{V}^{'}\|_{\text{QLoRa}}}, \quad \mathbf{U}_{\text{opt}} = \frac{\mathbf{U}^{'}}{\|\mathbf{U}^{'}\|_{\text{QLoRa}}},
$$
$$
\hat{\mathbf{V}}_{\text{quant}} = s \cdot \text{quantize}(\mathbf{V}_{\text{opt}} \cdot \mathbf{U}_{\text{opt}}^T), \quad \hat{\mathbf{U}}_{\text{quant}} = \hat{\mathbf{V}}_{\text{quant}}^T
$$

Here, \(\|\cdot\|_{\text{QLoRa}}\) denotes the normalization process enhanced by QLoRa, and \(\text{quantize}(\cdot)\) applies quantization for reduced model size and computational efficiency.

\subsubsection{Loss Computation with Enhanced Data Diversity}
We compute the vision and text prediction loss against a diversified and expanded ground truth vector \(\hat{z}\) so that it reflects the inclusion of synthetic data:
% \vspace{-2cm}
$$
\ell_{\text{vis}} = \ell(\hat{\mathbf{V}}_{\text{quant}}, \hat{z}_{\textit{st}}), \quad \ell_{\text{text}} = \ell(\hat{\mathbf{U}}_{\text{quant}}, \hat{z}_{\textit{st}})
$$

\subsubsection{FL Optimization with Adaptive Parameters}
We optimize the FL process by aggregating the compressed adaptive parameters \(w^h_{\text{opt}}\), in the FL server that eventually minimize the computational burden and reduces the communication overhead. We can present this process as follows:
% Optimizing for FL involves aggregating the efficiently compressed adapter parameters, \(w^h_{\text{opt}}\), across clients, minimizing both computational and communication overhead:
$$
w^{h, \text{opt}}_{\text{final}} = \sum_{i=1}^N \frac{m_i}{\sum_{j=1}^N m_j} \text{QLoRa}(\text{quantize}(w_i^h))
$$
This step leverages the advantages of quantization and QLoRa that ensures effective communication and efficient learning across the FL environment.

% This final step incorporates the benefits of both QLoRa and quantization to ensure efficient learning and communication in the FL environment, leveraging the comprehensive feature set enhanced by synthetic data generation for a robust, adaptable, and efficient FL framework.

In Fig. \ref{fig:1}, we present the overview of the FL framework, TriplePlay. At the beginning of our process, a pretrained CLIP model (which is possessed by each FL client) extracts task-specific features from the visual and textual data (denoted as I and T in Fig. \ref{fig:1}). Subsequently, an adaptor is tailored for each FL client and each client undergoes local training on their local available dataset. We fine-tune the adapter applying QLoRa that integrates both quantization and low-rank adaptation that compress the model size and minimize the resource consumption. Each FL client then share the refined local model with the server. On the server side, all the local models parameters are aggregated through weighted average to produce an updated global model which was randomly initialized at the beginning of the training process. The updated global model is distributed among all the participated FL clients so that each client can learn from the global model and update the parameters of its adapter. This iterative cycle of local model training, share local model with the server, aggregation and finally, redistribution of the global model is continued until the global model reaches a convergence.

% This cycle of local training, uploading, aggregation, and redistribution continues iteratively, aiming for convergence or until a predetermined number of rounds is completed.

\begin{algorithm}
\caption{\textit{TriplePlay:} Enhanced Federated Learning with CLIP and Synthetic Data}
\begin{algorithmic}[1]
\State \textbf{Input:} Set of input-label pairs $\{(\mathbf{v}_i, z_i)\}$ for $i=1$ to $N$, where $N$ is the number of clients
\State \textbf{Output:} Optimized global model parameters $w^{h, \text{opt}}_{\text{final}}$

\For{each client $i = 1$ to $N$}
    \State Generate synthetic data using GANs for $(\mathbf{v}_i, z_i)$
    \State Extract features using pretrained CLIP model:
     $\mathbf{V}_{\textit{st}} = f_{\text{vis}}(\text{GAN}(\mathbf{v}_i)), \mathbf{U}_{\textit{st}} = f_{\text{text}}(\text{GAN}(z_i))$
    \State Apply adaptive refinement mechanism $h$ to enhance task-specific features:
    \State \quad $\mathbf{V}^{'} = h(\mathbf{V}_{\textit{st}}) \otimes \mathbf{V}_{\textit{st}}$
    \State \quad $\mathbf{U}^{'} = h(\mathbf{U}_{\textit{st}}) \otimes \mathbf{U}_{\textit{st}}$
    \State Normalize and interact features with QLoRa and quantization optimization:
    \State \quad $\mathbf{V}_{\text{opt}} = \frac{\mathbf{V}^{'}}{\|\mathbf{V}^{'}\|_{\text{QLoRa}}}, \mathbf{U}_{\text{opt}} = \frac{\mathbf{U}^{'}}{\|\mathbf{U}^{'}\|_{\text{QLoRa}}}$
    \State \quad $\hat{\mathbf{V}}_{\text{quant}} = s \cdot \text{quantize}(\mathbf{V}_{\text{opt}} \cdot \mathbf{U}_{\text{opt}}^T), \hat{\mathbf{U}}_{\text{quant}} = \hat{\mathbf{V}}_{\text{quant}}^T$
    \State Compute loss with the enhanced data diversity:
    \State \quad $\ell_{\text{vis}} = \ell(\hat{\mathbf{V}}_{\text{quant}}, \hat{z}_{\textit{st}}), \ell_{\text{text}} = \ell(\hat{\mathbf{U}}_{\text{quant}}, \hat{z}_{\textit{st}})$
    \State Update local model parameters based on the computed loss
\EndFor

\State Aggregate adaptive parameters across clients with QLoRa and quantization for global model update:
\State \quad $w^{h, \text{opt}}_{\text{final}} = \sum_{i=1}^N \frac{m_i}{\sum_{j=1}^N m_j} \text{QLoRa}(\text{quantize}(w_i^h))$

\State \textbf{return} the optimized global model parameters $w^{h, \text{opt}}_{\text{final}}$
\end{algorithmic}
\end{algorithm}

\begin{figure*}[htb!] % Defines figure environment
    \centering % Centers your figure
    \includegraphics[width=0.97\linewidth]{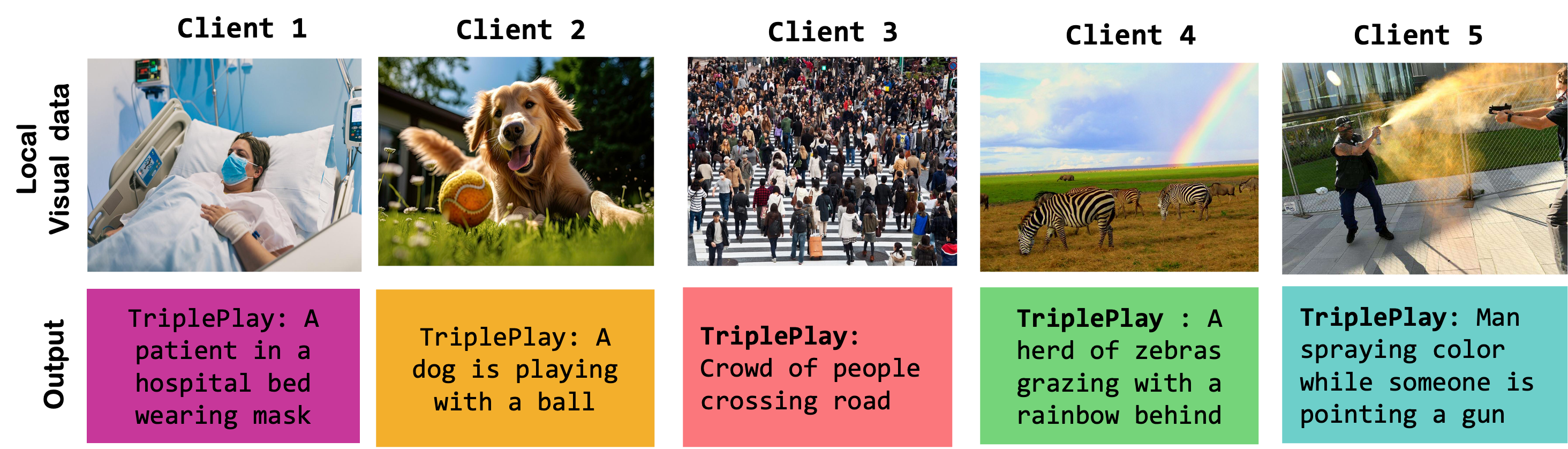} % Includes your figure and defines the size
    % \vspace{-0.4cm}
    \caption{Image showcases the visual data from five different FL clients and the corresponding outputs generated by the TriplePlay.}  % For your caption
    \label{fig:TriplePlay-output}
\end{figure*}

% \begin{algorithm}
% \caption{Training procedure for a basic GAN}
% \begin{algorithmic}[1]
% \State Initialize generator $G$ and discriminator $D$ with random weights
% \For{each training iteration}
%     \State Sample minibatch of $m$ noise samples $\{z^{(1)}, ..., z^{(m)}\}$ from noise prior $p_g(z)$
%     \State Sample minibatch of $m$ examples $\{x^{(1)}, ..., x^{(m)}\}$ from data generating distribution $p_{data}(x)$
%     \State Update the discriminator by ascending its stochastic gradient:
%     \[
%     \nabla_{\theta_d} \frac{1}{m} \sum_{i=1}^m [\log D(x^{(i)}) + \log (1 - D(G(z^{(i)})))]
%     \]
%     \State Sample minibatch of $m$ noise samples $\{z^{(1)}, ..., z^{(m)}\}$ from noise prior $p_g(z)$
%     \State Update the generator by descending its stochastic gradient:
%     \[
%     \nabla_{\theta_g} \frac{1}{m} \sum_{i=1}^m \log (1 - D(G(z^{(i)})))
%     \]
% \EndFor
% \end{algorithmic}
% \end{algorithm}

\section{Experimental Analysis}
% We conduct experiments in two phases to assess the effectiveness and scalability of our approach, TriplePlay. In the first phase, we experiment with 5 clients to evaluate the impact of QLoRa fine-tuning on performance and GPU utilization in a multi-client scenario. 
% In the second phase, we extend our experiments to 10 clients to demonstrate the scalability of TriplePlay. 

\subsection{Dataset}
We evaluate our proposed approach on PACS \cite{li2017deeper} and Office-Home \cite{venkateswara2017deep} dataset. PACS consists of four domains (photo, art painting, cartoon, and sketch) with a total of $9,991$ images distributed across $7$ object categories. On the other hand, Office-Home \cite{venkateswara2017deep} is a classification benchmark dataset, containing approximately $15,500$ images across $65$ classes. Notably, in the PACS dataset, the `Photo' class exhibits fewer samples than other classes, while in the Office-Home dataset, the `Product' class contains less amount of data samples in comparison to other classes. Hence, we apply GAN to generate synthetic data for these underrepresented classes.

\subsection{Result Analysis}
\begin{figure}[htb!]
 \setlength{\belowcaptionskip}{-10pt}
\centering
\begin{tabular}{cc}
\includegraphics[width=0.47\linewidth]{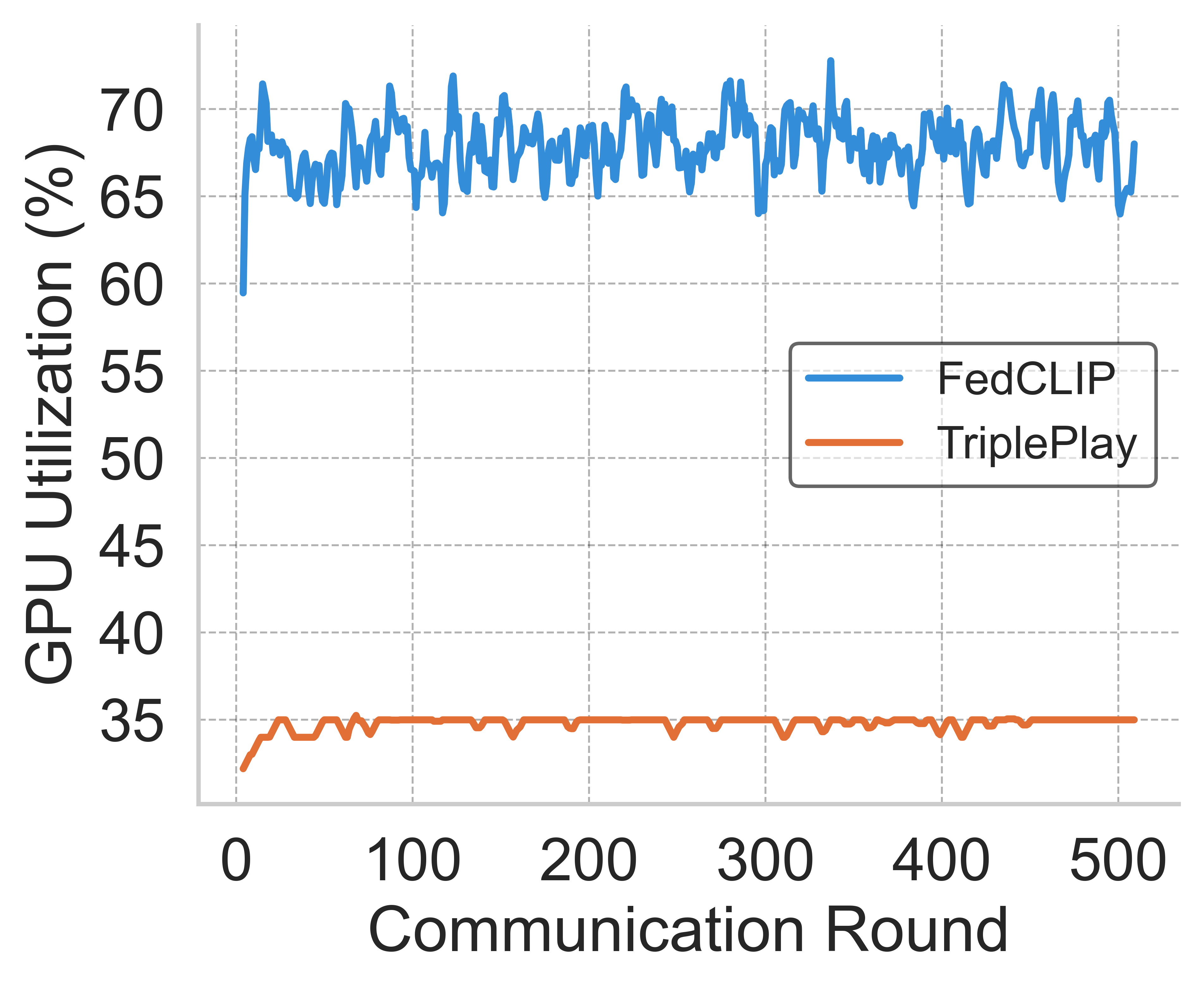}
&
\includegraphics[width=0.47\linewidth]{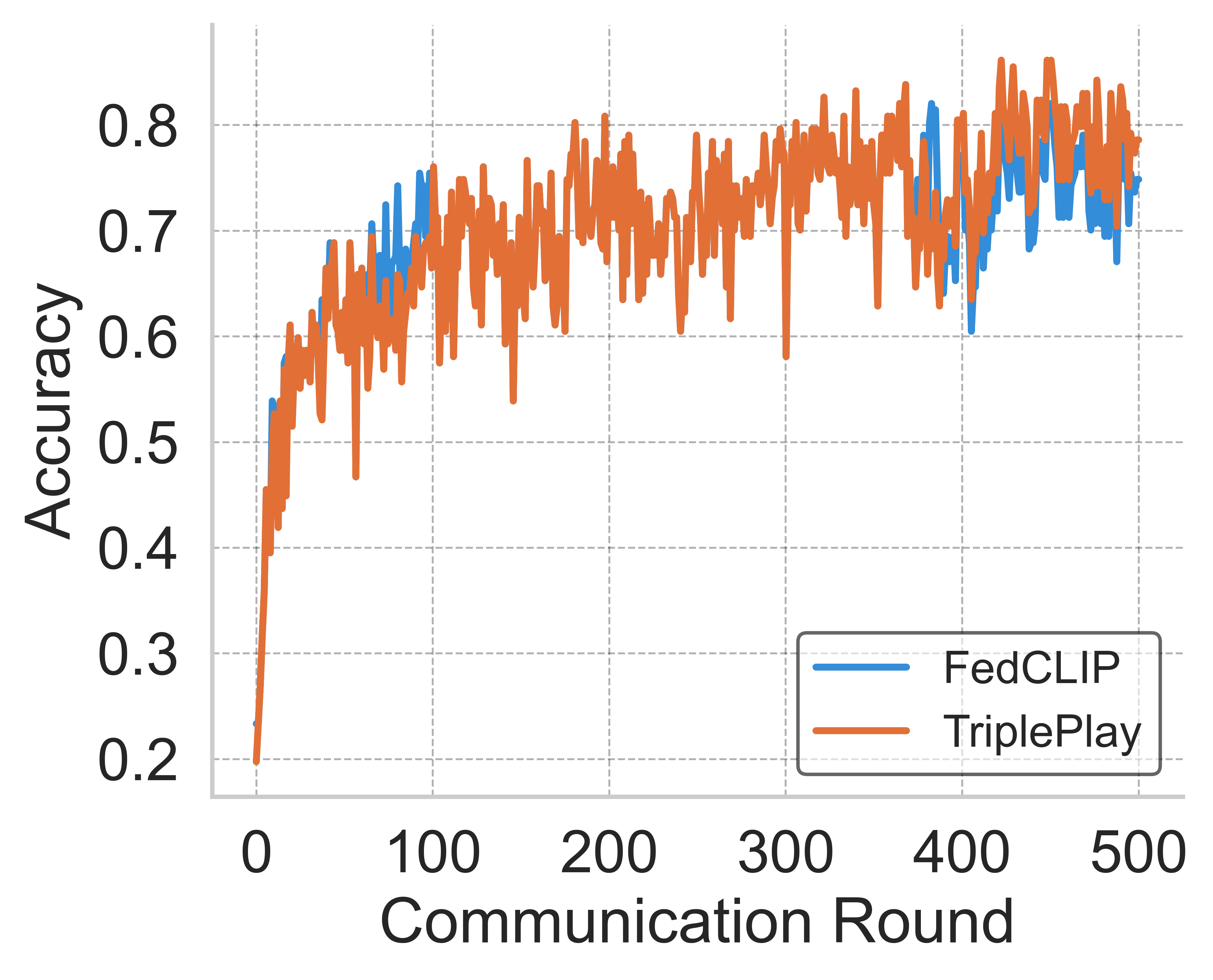}
\end{tabular}
\caption{GPU Utilization (left) and Test Accuracy (right) in Vanilla FedCLIP and our approach with  on PACS Dataset.
}
\label{fig:gpuvsacc}
\end{figure}

In Fig. \ref{fig:TriplePlay-output}, we demonstrate the TriplePlay's 
proficiency in semantic understanding and precise text generation tailored to the specific visual context of each client. Each FL client's local visual data, ranging from the activity of a hospital patient to a zebra herd, is processed to produce accurate and contextually appropriate textual descriptions. This highlights the model's ability to understand and articulate fine-grained details from diverse visual inputs, ensuring personalized and relevant textual outputs based on the local datasets. 
In Figure \ref{fig:gpuvsacc} (left), the resource usage of FedCLIP versus TriplePlay is illustrated through a line graph that details the percentage of GPU utilization across a spectrum of communication rounds from $0$ to $500$.  The FedCLIP line exhibits significant fluctuations, with utilization percentages oscillating between approximately 60\% and 70\%, indicating a variable demand on the GPU resources throughout the communication rounds. In stark contrast, our proposed approach, TriplePlay maintains a remarkably steady and lower GPU utilization, consistently around the 35\% mark, indicating a more efficient and stable usage of GPU resources over time. This visual data clearly suggests that the TriplePlay method ensures a more uniform and possibly more efficient GPU usage profile compared to FedCLIP.

In Figure \ref{fig:gpuvsacc} (right), we present a comparison of the accuracy trajectories between the FedCLIP methodology and TriplePlay over $500$ communication epochs. FedCLIP improves gradually with a starting accuracy below $60\%$, and shows a steady model improvement. Conversely, our proposed method, TriplePlay demonstrates a sharp increase in accuracy early on, crossing the $0.6$ threshold within the initial $50$ rounds. After this quick rise, while the accuracy continues to improve, it does so at a more moderate pace, surpassing the $0.7$ level just beyond the 100th round. 
% \vspace{-0.25cm}

\begin{figure}[htb!]
%\setlength{\belowcaptionskip}{-15pt}
%\vspace{-0.25cm}
\centering
%\footnotesize
\includegraphics[width=0.9\linewidth]{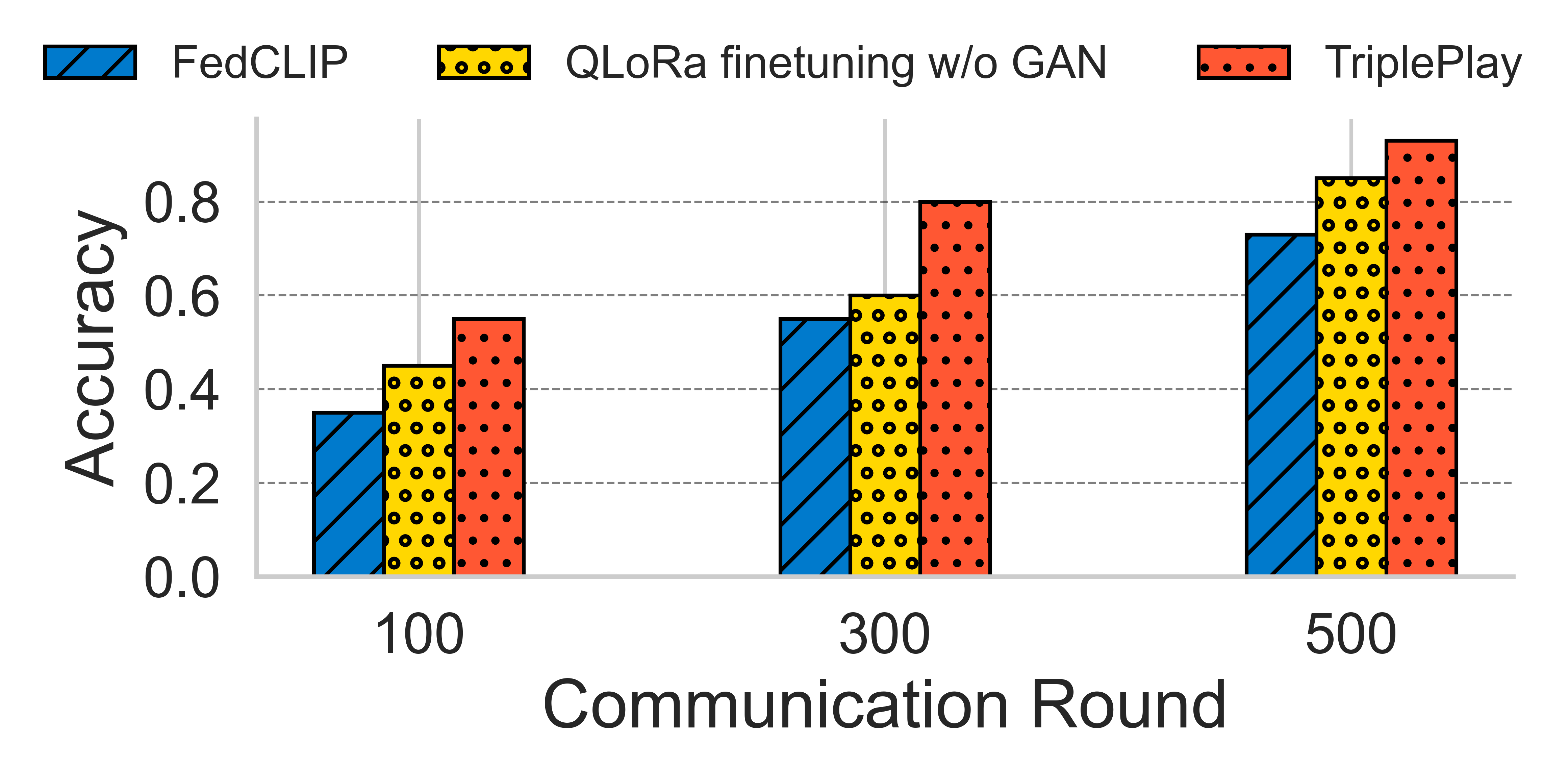}
% \vspace{-0.25cm}
\caption{Server accuracy comparison among FedCLIP, QLora fine-tuning without GAN, and TriplePlay on PACS dataset. 
}
\label{fig:baracc}
\end{figure}

\begin{figure}[h]
 %\setlength{\belowcaptionskip}{-10pt}
% \vspace{-0.25cm}
\centering
%\footnotesize
\includegraphics[width=0.9\linewidth]{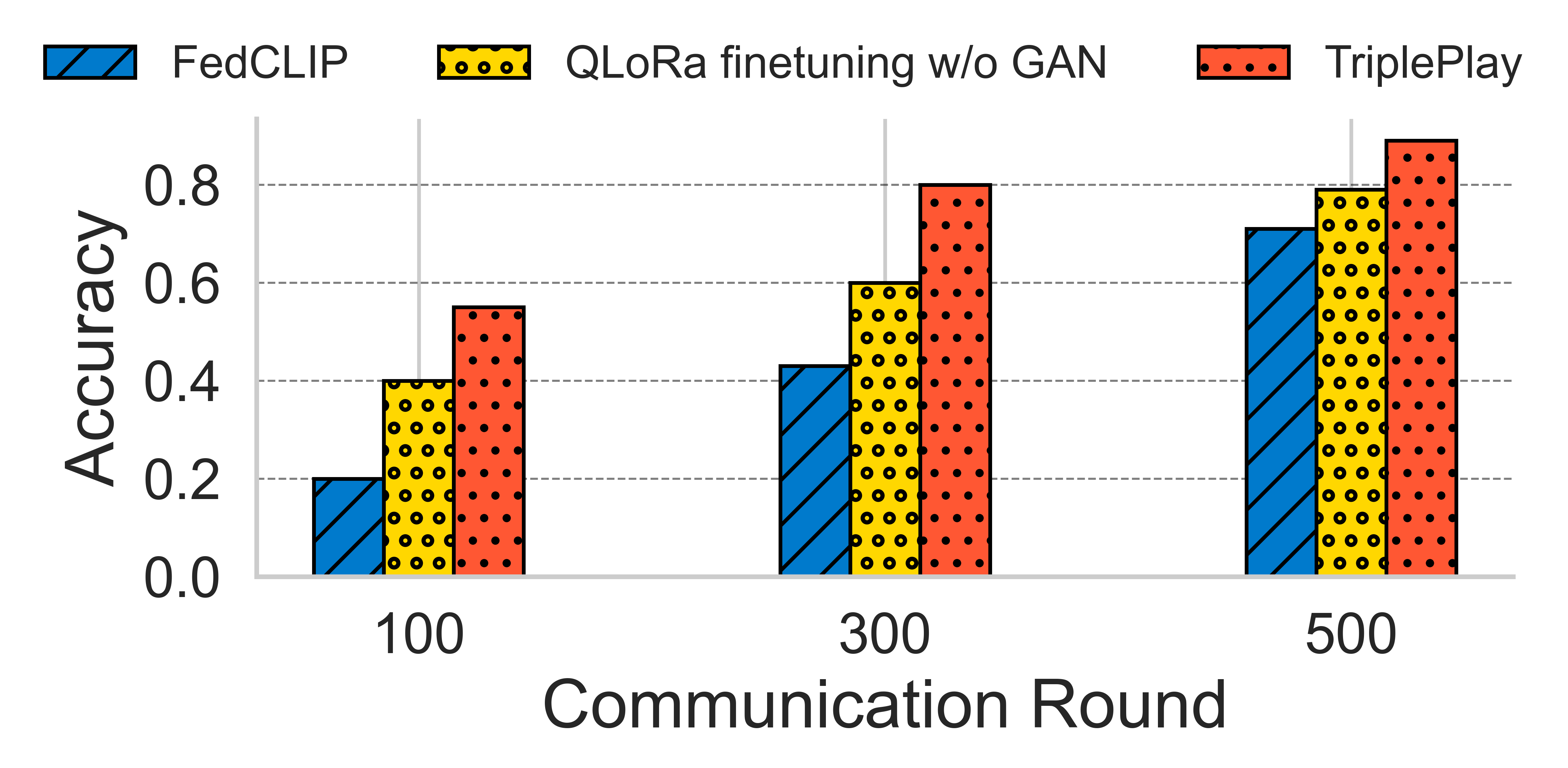}
% \vspace{-0.2cm}
\caption{Server accuracy comparison among Vanilla FedCLIP, FedCLIP with QLora, and TriplePlay on Office-Home dataset. 
}
\label{fig:baraccoffice}
\end{figure}

\begin{figure}[htb!]
%\setlength{\belowcaptionskip}{-10pt}
% \vspace{-0.16cm}
\centering
%\footnotesize
\begin{tabular}{cc}
\includegraphics[width=0.46\linewidth]{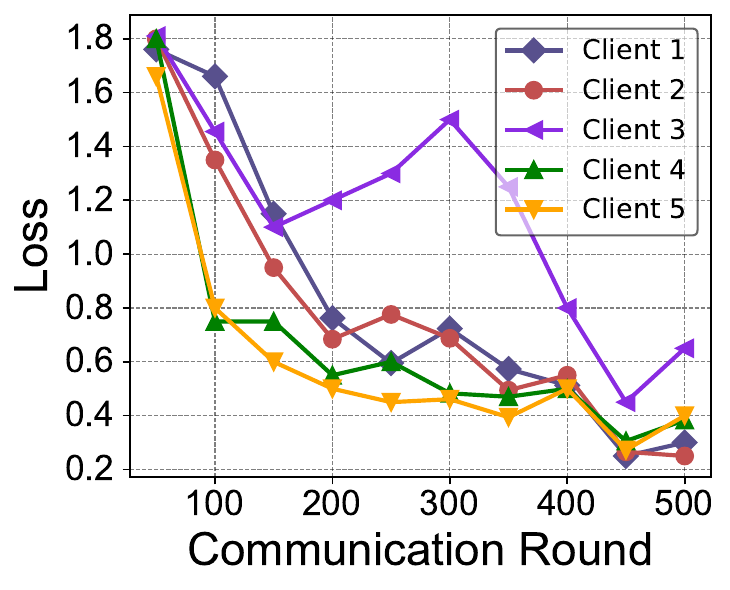}
&
\includegraphics[width=0.46\linewidth]{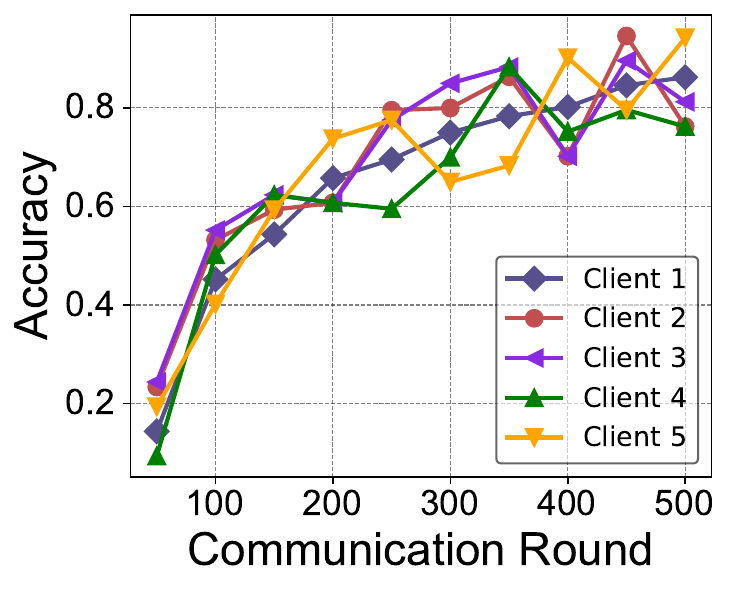}
\\
 % (a) & (b) \\
\end{tabular}
% \vspace{-0.29cm}
\caption{Loss (left) and accuracy (right) of individual clients with TriplePlay on the PACS dataset.}
\label{fig:clients_res}
\end{figure}

To demonstrate a clear performance comparison among FedCLIP, QLoRA method without fine-tuning and TriplePlay, we presented the accuracy of each of these approach in Figure \ref{fig:baracc} considering PACS dataset. We also evaluated the performance considering Office-Home dataset (Fig. \ref{fig:baraccoffice}). From the performance comparisons in Figures \ref{fig:baracc} and \ref{fig:baraccoffice}, it is clear that TriplePlay outperforms the other two methods.
Since TriplePlay tailors GANs to counteract the impact of class imbalance, it generalizes faster compare to other methods. This results in significant performance gains, with TriplaPlay achieves $80\%$ accuracy within just $300$ communication rounds. While the QLoRA fine-tuning method shows better results compared to vanilla FedCLIP, it fails to achieve the desired accuracy because of the class imbalance issue.
% \vspace{-0.25cm}
In Figure \ref{fig:clients_res}, we illustrate individual client's loss minimization and accuracy over 500 communication rounds. We observe that in terms of loss minimization, each client exhibits a consistent decrease in loss as the number of communication rounds increases. This indicates that the TriplePlay effectively minimizes the loss function for each client participating in the FL process. Similarly, the accuracy graph demonstrates a steady improvement in performance across all clients. The increasing accuracy values suggest that TriplePlay successfully learns and generalizes from the distributed data. 

\begin{figure}[htb!]
\vspace{-0.16cm}
\centering
%\footnotesize
\begin{tabular}{cc}
\includegraphics[width=0.46\linewidth]{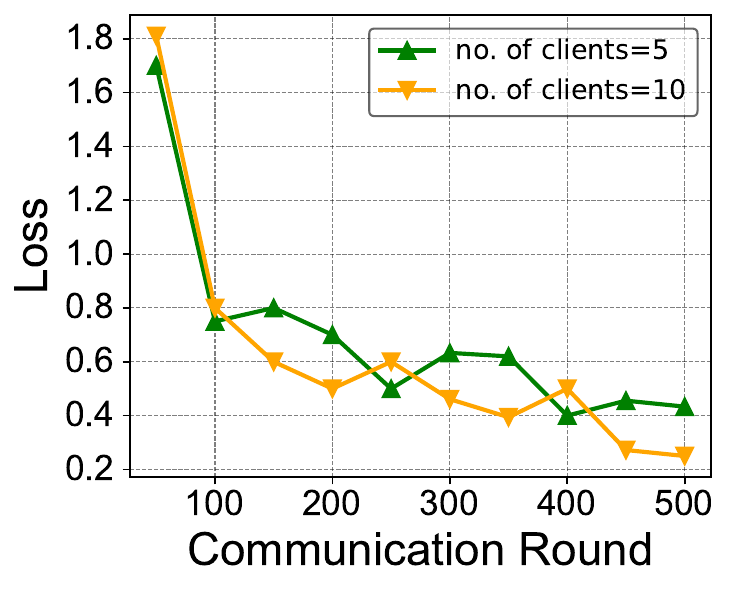}
&
\includegraphics[width=0.46\linewidth]{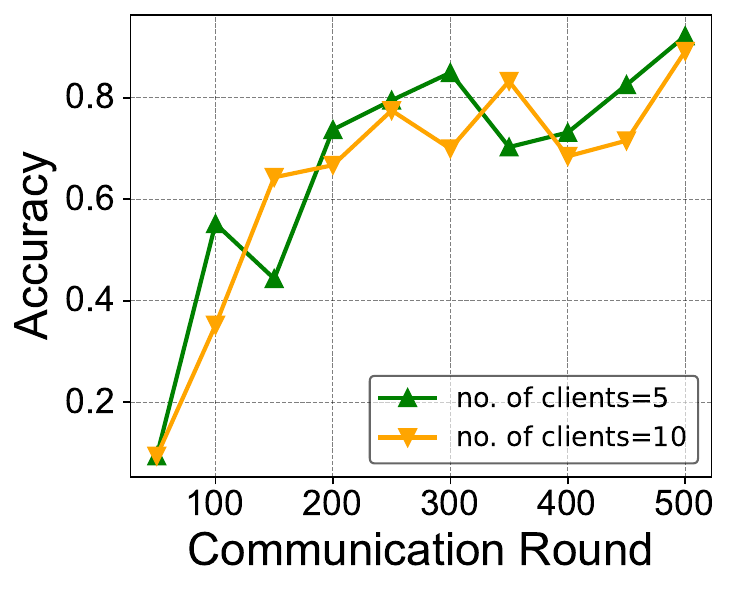}
\\
 % (a) & (b) \\
\end{tabular}
% \vspace{-0.2cm}
\caption{Server loss (left) and accuracy (right) of TriplePlay with varying number of clients on PACS dataset.}
\label{fig:server_res}
\end{figure}

We further evaluated the scalability of TriplePlay considering higher number of FL clients within the network and showcases our results 
in Fig. \ref{fig:server_res}. For this experiment, in this paper, we presented evaluations considering $5$ and $10$ FL network client configurations over $500$ communication rounds. For 5-FL client configuration, the FL global model loss decreases in a steady manner and the corresponding accuracy increases. After that, we present the impact while considering $10$ FL clients, and we observe a similar trend: a consistent loss minimization and accuracy improvement of the FL clients' local models. However, in the 10-client setting, we notice a slight improvement of the local model accuracy of the FL clients. We also evaluated the performance of TriplePlay in various client settings, and it shows significant performance improvement over a range of communication rounds.  These outcomes shows the potential of TriplePlay in terms of scalability. 

% These findings underscore the scalability of TriplePlay, as it maintains its effectiveness in terms of loss minimization and accuracy enhancement even when the number of clients is doubled.

\section{Conclusion}

In this paper, we presented how large foundation model (e.g., CLIP) can be integrated in an FL-based system considering the challenges of resource constraints and heterogeneous data distributions in the network clients. With our proposed approach, we achieved improved performance in presence of heterogeneous data distributions among the clients, specially with minority classes, and marginal reduction in the communication round and computational resources. This research will pave the way to conduct privacy-preserving and efficient model training in a distributed learning settings.

% This paper has set forth a framework to address the pressing challenges in FL, especially in the context of integrating large foundation models like CLIP and managing data diversity and resource constraints. The outcomes include improved adaptability of FL systems across varied data distributions, enhanced model performance, especially for minority classes, and a reduction in the computational and communication resources required. Ultimately, this research aims to push the boundaries of what is currently achievable in FL, paving the way for more inclusive, efficient, and privacy-preserving machine learning models that can adapt to and thrive within the complex, data-diverse landscapes of the real world.

% \section{Acknowledgement}
% This material is based upon work supported by the National Science Foundation (NSF) under Grant No. CRII-IIS-RI-2348145. Any opinions, findings, and conclusions or recommendations expressed in this material are those of the author(s) and do not necessarily reflect the views of the NSF.

 % \section{Acknowledgement}
 % This material is based upon work supported by the National Science Foundation (NSF) under Grant No. XXX-XXX-XXXXXXX.
 % %CRII-IIS-RI-2348145
 % Any opinions, findings, and conclusions or recommendations expressed in this material are those of the author(s) and do not necessarily reflect the views of the NSF.

\bibliographystyle{IEEEtran}
\bibliography{ref.bib}

\begin{thebibliography}{10}
\providecommand{\url}[1]{#1}
\csname url@samestyle\endcsname
\providecommand{\newblock}{\relax}
\providecommand{\bibinfo}[2]{#2}
\providecommand{\BIBentrySTDinterwordspacing}{\spaceskip=0pt\relax}
\providecommand{\BIBentryALTinterwordstretchfactor}{4}
\providecommand{\BIBentryALTinterwordspacing}{\spaceskip=\fontdimen2\font plus
\BIBentryALTinterwordstretchfactor\fontdimen3\font minus \fontdimen4\font\relax}
\providecommand{\BIBforeignlanguage}[2]{{%
\expandafter\ifx\csname l@#1\endcsname\relax
\typeout{** WARNING: IEEEtran.bst: No hyphenation pattern has been}%
\typeout{** loaded for the language `#1'. Using the pattern for}%
\typeout{** the default language instead.}%
\else
\language=\csname l@#1\endcsname
\fi
#2}}
\providecommand{\BIBdecl}{\relax}
\BIBdecl

\bibitem{radford2021learning}
A.~Radford, J.~W. Kim, C.~Hallacy, A.~Ramesh, G.~Goh, S.~Agarwal, G.~Sastry, A.~Askell, P.~Mishkin, J.~Clark \emph{et~al.}, ``Learning transferable visual models from natural language supervision,'' in \emph{International conference on machine learning}.\hskip 1em plus 0.5em minus 0.4em\relax PMLR, 2021, pp. 8748--8763.

\bibitem{mcmahan2017communication}
B.~McMahan, E.~Moore, D.~Ramage, S.~Hampson, and B.~A. y~Arcas, ``Communication-efficient learning of deep networks from decentralized data,'' in \emph{Artificial intelligence and statistics}.\hskip 1em plus 0.5em minus 0.4em\relax PMLR, 2017, pp. 1273--1282.

\bibitem{nguyen2021federated}
D.~C. Nguyen, M.~Ding, P.~N. Pathirana, A.~Seneviratne, J.~Li, and H.~V. Poor, ``Federated learning for internet of things: A comprehensive survey,'' \emph{IEEE Communications Surveys \& Tutorials}, vol.~23, no.~3, pp. 1622--1658, 2021.

\bibitem{imteaj2021survey}
A.~Imteaj, U.~Thakker, S.~Wang, J.~Li, and M.~H. Amini, ``A survey on federated learning for resource-constrained iot devices,'' \emph{IEEE Internet of Things Journal}, vol.~9, no.~1, pp. 1--24, 2021.

\bibitem{hossain2024flamingo}
M.~Z. Hossain, A.~Imteaj, and A.~R. Shahid, ``Flamingo: Adaptive and resilient federated meta-learning against adversarial attacks,'' in \emph{2024 IEEE 44th International Conference on Distributed Computing Systems Workshops (ICDCSW)}.\hskip 1em plus 0.5em minus 0.4em\relax IEEE, 2024, pp. 17--22.

\bibitem{9356216}
A.~Imteaj and M.~Hadi~Amini, ``Fedar: Activity and resource-aware federated learning model for distributed mobile robots,'' in \emph{2020 19th IEEE International Conference on Machine Learning and Applications (ICMLA)}, 2020, pp. 1153--1160.

\bibitem{chen2022bridging}
H.-Y. Chen and W.-L. Chao, ``On bridging generic and personalized federated learning for image classification,'' in \emph{International Conference on Learning Representations}, 2022.

\bibitem{gupta2022fl}
S.~Gupta, K.~Ahuja, M.~Havaei, N.~Chatterjee, and Y.~Bengio, ``Fl games: A federated learning framework for distribution shifts,'' in \emph{Workshop on Federated Learning: Recent Advances and New Challenges (in Conjunction with NeurIPS 2022)}, 2022.

\bibitem{qu2022generalized}
Z.~Qu, X.~Li, R.~Duan, Y.~Liu, B.~Tang, and Z.~Lu, ``Generalized federated learning via sharpness aware minimization,'' in \emph{International conference on machine learning}.\hskip 1em plus 0.5em minus 0.4em\relax PMLR, 2022, pp. 18\,250--18\,280.

\bibitem{hossain2024fedavo}
M.~Z. Hossain, A.~Imteaj, and A.~R. Shahid, ``Fedavo: Improving communication efficiency in federated learning with african vultures optimizer,'' in \emph{2024 IEEE 48th Annual Computers, Software, and Applications Conference (COMPSAC)}.\hskip 1em plus 0.5em minus 0.4em\relax IEEE, 2024, pp. 455--462.

\bibitem{foret2020sharpness}
P.~Foret, A.~Kleiner, H.~Mobahi, and B.~Neyshabur, ``Sharpness-aware minimization for efficiently improving generalization,'' in \emph{International Conference on Learning Representations}, 2020.

\bibitem{li2021ditto}
T.~Li, S.~Hu, A.~Beirami, and V.~Smith, ``Ditto: Fair and robust federated learning through personalization,'' in \emph{International conference on machine learning}.\hskip 1em plus 0.5em minus 0.4em\relax PMLR, 2021, pp. 6357--6368.

\bibitem{tan2022towards}
A.~Z. Tan, H.~Yu, L.~Cui, and Q.~Yang, ``Towards personalized federated learning,'' \emph{IEEE Transactions on Neural Networks and Learning Systems}, 2022.

\bibitem{tan2022federated}
Y.~Tan, G.~Long, J.~Ma, L.~Liu, T.~Zhou, and J.~Jiang, ``Federated learning from pre-trained models: A contrastive learning approach,'' \emph{Advances in neural information processing systems}, vol.~35, pp. 19\,332--19\,344, 2022.

\bibitem{tian2022fedbert}
Y.~Tian, Y.~Wan, L.~Lyu, D.~Yao, H.~Jin, and L.~Sun, ``Fedbert: When federated learning meets pre-training,'' \emph{ACM Transactions on Intelligent Systems and Technology (TIST)}, vol.~13, no.~4, pp. 1--26, 2022.

\bibitem{chen2022importance}
H.-Y. Chen, C.-H. Tu, Z.~Li, H.-W. Shen, and W.-L. Chao, ``On the importance and applicability of pre-training for federated learning,'' \emph{arXiv preprint arXiv:2206.11488}, 2022.

\bibitem{guo2023promptfl}
T.~Guo, S.~Guo, J.~Wang, X.~Tang, and W.~Xu, ``Promptfl: Let federated participants cooperatively learn prompts instead of models-federated learning in age of foundation model,'' \emph{IEEE Transactions on Mobile Computing}, 2023.

\bibitem{jia2021scaling}
C.~Jia, Y.~Yang, Y.~Xia, Y.-T. Chen, Z.~Parekh, H.~Pham, Q.~Le, Y.-H. Sung, Z.~Li, and T.~Duerig, ``Scaling up visual and vision-language representation learning with noisy text supervision,'' in \emph{International conference on machine learning}.\hskip 1em plus 0.5em minus 0.4em\relax PMLR, 2021, pp. 4904--4916.

\bibitem{li2022blip}
J.~Li, D.~Li, C.~Xiong, and S.~Hoi, ``Blip: Bootstrapping language-image pre-training for unified vision-language understanding and generation,'' in \emph{International conference on machine learning}.\hskip 1em plus 0.5em minus 0.4em\relax PMLR, 2022, pp. 12\,888--12\,900.

\bibitem{singh2022flava}
A.~Singh, R.~Hu, V.~Goswami, G.~Couairon, W.~Galuba, M.~Rohrbach, and D.~Kiela, ``Flava: A foundational language and vision alignment model,'' in \emph{Proceedings of the IEEE/CVF Conference on Computer Vision and Pattern Recognition}, 2022, pp. 15\,638--15\,650.

\bibitem{wang2021simvlm}
Z.~Wang, J.~Yu, A.~W. Yu, Z.~Dai, Y.~Tsvetkov, and Y.~Cao, ``Simvlm: Simple visual language model pretraining with weak supervision,'' in \emph{International Conference on Learning Representations}, 2021.

\bibitem{chen2023altclip}
Z.~Chen, G.~Liu, B.-W. Zhang, Q.~Yang, and L.~Y. Wu, ``Altclip: Altering the language encoder in clip for extended language capabilities,'' in \emph{The 61st Annual Meeting Of The Association For Computational Linguistics}, 2023.

\bibitem{chia2022contrastive}
P.~J. Chia, G.~Attanasio, F.~Bianchi, S.~Terragni, A.~R. Magalh{\~a}es, D.~Goncalves, C.~Greco, and J.~Tagliabue, ``Contrastive language and vision learning of general fashion concepts,'' \emph{Scientific Reports}, vol.~12, no.~1, p. 18958, 2022.

\bibitem{huang2023visual}
Z.~Huang, F.~Bianchi, M.~Yuksekgonul, T.~J. Montine, and J.~Zou, ``A visual--language foundation model for pathology image analysis using medical twitter,'' \emph{Nature medicine}, vol.~29, no.~9, pp. 2307--2316, 2023.

\bibitem{wang2023fashionvqa}
M.~Wang, A.~Mahjoubfar, and A.~Joshi, ``Fashionvqa: A domain-specific visual question answering system,'' in \emph{Proceedings of the IEEE/CVF Conference on Computer Vision and Pattern Recognition}, 2023, pp. 3513--3518.

\bibitem{du2024domain}
Z.~Du, X.~Li, F.~Li, K.~Lu, L.~Zhu, and J.~Li, ``Domain-agnostic mutual prompting for unsupervised domain adaptation,'' in \emph{Proceedings of the IEEE/CVF Conference on Computer Vision and Pattern Recognition}, 2024, pp. 23\,375--23\,384.

\bibitem{wang2024grammar}
B.~Wang, Z.~Wang, X.~Wang, Y.~Cao, R.~A~Saurous, and Y.~Kim, ``Grammar prompting for domain-specific language generation with large language models,'' \emph{Advances in Neural Information Processing Systems}, vol.~36, 2024.

\bibitem{lian2024recai}
J.~Lian, Y.~Lei, X.~Huang, J.~Yao, W.~Xu, and X.~Xie, ``Recai: Leveraging large language models for next-generation recommender systems,'' in \emph{Companion Proceedings of the ACM on Web Conference 2024}, 2024, pp. 1031--1034.

\bibitem{gouidis2024fusing}
F.~Gouidis, K.~Papantoniou, K.~Papoutsakis, T.~Patkos, A.~Argyros, and D.~Plexousakis, ``Fusing domain-specific content from large language models into knowledge graphs for enhanced zero shot object state classification,'' in \emph{Proceedings of the AAAI Symposium Series}, vol.~3, no.~1, 2024, pp. 115--124.

\bibitem{li2023igg}
P.~Li, Y.~He, F.~R. Yu, P.~Song, D.~Yin, and G.~Zhou, ``Igg: Improved graph generation for domain adaptive object detection,'' in \emph{Proceedings of the 31st ACM International Conference on Multimedia}, 2023, pp. 1314--1324.

\bibitem{sun2024training}
W.~Sun, Y.~Du, G.~Liu, R.~Kompella, and C.~G. Snoek, ``Training-free semantic segmentation via llm-supervision,'' \emph{arXiv preprint arXiv:2404.00701}, 2024.

\bibitem{wang2024llm}
J.~Wang and L.~Ke, ``Llm-seg: Bridging image segmentation and large language model reasoning,'' \emph{arXiv preprint arXiv:2404.08767}, 2024.

\bibitem{wei2024lasagna}
C.~Wei, H.~Tan, Y.~Zhong, Y.~Yang, and L.~Ma, ``Lasagna: Language-based segmentation assistant for complex queries,'' \emph{arXiv preprint arXiv:2404.08506}, 2024.

\bibitem{cha2023learning}
J.~Cha, J.~Mun, and B.~Roh, ``Learning to generate text-grounded mask for open-world semantic segmentation from only image-text pairs,'' in \emph{Proceedings of the IEEE/CVF Conference on Computer Vision and Pattern Recognition}, 2023, pp. 11\,165--11\,174.

\bibitem{li2023object}
Z.~Li, P.~Xu, X.~Chang, L.~Yang, Y.~Zhang, L.~Yao, and X.~Chen, ``When object detection meets knowledge distillation: A survey,'' \emph{IEEE Transactions on Pattern Analysis and Machine Intelligence}, 2023.

\bibitem{park2024localized}
J.~S. Park, J.~Hessel, K.~Chandu, P.~P. Liang, X.~Lu, P.~West, Y.~Yu, Q.~Huang, J.~Gao, A.~Farhadi \emph{et~al.}, ``Localized symbolic knowledge distillation for visual commonsense models,'' \emph{Advances in Neural Information Processing Systems}, vol.~36, 2024.

\bibitem{10633382}
S.~Zaman, S.~Talukder, M.~Z. Hossain, S.~M.~T. Puppala, and A.~Imteaj, ``Towards communication-efficient federated learning through particle swarm optimization and knowledge distillation,'' in \emph{2024 IEEE 48th Annual Computers, Software, and Applications Conference (COMPSAC)}, 2024, pp. 510--518.

\bibitem{kang2024knowledge}
M.~Kang, S.~Lee, J.~Baek, K.~Kawaguchi, and S.~J. Hwang, ``Knowledge-augmented reasoning distillation for small language models in knowledge-intensive tasks,'' \emph{Advances in Neural Information Processing Systems}, vol.~36, 2024.

\bibitem{imteaj2023fedmdp}
A.~Imteaj and M.~Amini, ``Fedmdp: A federated learning framework to handle system and model heterogeneity in resource-constrained environments,'' in \emph{Proc. AAAI Conf. Artif. Intell}, 2023.

\bibitem{hossain2024securing}
M.~Z. Hossain and A.~Imteaj, ``Securing vision-language models with a robust encoder against jailbreak and adversarial attacks,'' \emph{arXiv preprint arXiv:2409.07353}, 2024.

\bibitem{hossain2024sim}
------, ``Sim-clip: Unsupervised siamese adversarial fine-tuning for robust and semantically-rich vision-language models,'' \emph{arXiv preprint arXiv:2407.14971}, 2024.

\bibitem{guo2023pfedprompt}
T.~Guo, S.~Guo, and J.~Wang, ``Pfedprompt: Learning personalized prompt for vision-language models in federated learning,'' in \emph{Proceedings of the ACM Web Conference 2023}, 2023, pp. 1364--1374.

\bibitem{yeh2023meta}
C.-H. Yeh, B.~Russell, J.~Sivic, F.~C. Heilbron, and S.~Jenni, ``Meta-personalizing vision-language models to find named instances in video,'' in \emph{Proceedings of the IEEE/CVF Conference on Computer Vision and Pattern Recognition}, 2023, pp. 19\,123--19\,132.

\bibitem{cho2023promptstyler}
J.~Cho, G.~Nam, S.~Kim, H.~Yang, and S.~Kwak, ``Promptstyler: Prompt-driven style generation for source-free domain generalization,'' in \emph{Proceedings of the IEEE/CVF International Conference on Computer Vision}, 2023, pp. 15\,702--15\,712.

\bibitem{yan2023clip}
S.~Yan, N.~Dong, L.~Zhang, and J.~Tang, ``Clip-driven fine-grained text-image person re-identification,'' \emph{IEEE Transactions on Image Processing}, 2023.

\bibitem{chen2024feddat}
H.~Chen, Y.~Zhang, D.~Krompass, J.~Gu, and V.~Tresp, ``Feddat: An approach for foundation model finetuning in multi-modal heterogeneous federated learning,'' in \emph{Proceedings of the AAAI Conference on Artificial Intelligence}, vol.~38, no.~10, 2024, pp. 11\,285--11\,293.

\bibitem{li2017deeper}
D.~Li, Y.~Yang, Y.-Z. Song, and T.~M. Hospedales, ``Deeper, broader and artier domain generalization,'' in \emph{Proceedings of the IEEE international conference on computer vision}, 2017, pp. 5542--5550.

\bibitem{venkateswara2017deep}
H.~Venkateswara, J.~Eusebio, S.~Chakraborty, and S.~Panchanathan, ``Deep hashing network for unsupervised domain adaptation,'' in \emph{Proceedings of the IEEE conference on computer vision and pattern recognition}, 2017, pp. 5018--5027.

\end{thebibliography}

\end{document}